\documentclass[conference]{IEEEtran}
\newcommand{\defense}{PATROL\@\xspace}
\IEEEoverridecommandlockouts
\usepackage[pagebackref,breaklinks,colorlinks]{hyperref}
\usepackage{cite}
\usepackage{amsfonts}
\usepackage{algorithmic}
\usepackage{graphicx}
\usepackage{amsmath}
\usepackage{amssymb}
\usepackage{booktabs}
\usepackage[linesnumbered,ruled]{algorithm2e}
\usepackage{textcomp}
\usepackage{xcolor}
\usepackage{float}
\usepackage{booktabs}
\usepackage{multirow}
\usepackage{enumerate}
\usepackage{bm}
\usepackage{subcaption}
\usepackage{CJKutf8}
\usepackage{url}
\usepackage{xcolor}
\usepackage{xspace}
\usepackage{enumitem}
\usepackage{txfonts}
\usepackage{makecell}
\usepackage{times}
\usepackage{epsfig}
\def\BibTeX{{\rm B\kern-.05em{\sc i\kern-.025em b}\kern-.08em
    T\kern-.1667em\lower.7ex\hbox{E}\kern-.125emX}}
\begin{document}

\title{\LARGE PATROL: \underline{P}riv\underline{a}cy-Orien\underline{t}ed P\underline{r}uning for C\underline{ol}laborative Inference Against \\Model Inversion Attacks}

\author{\IEEEauthorblockN{Shiwei Ding\IEEEauthorrefmark{2}, Lan Zhang\IEEEauthorrefmark{2}, Miao Pan\IEEEauthorrefmark{3}, Xiaoyong Yuan\IEEEauthorrefmark{2}}
\IEEEauthorblockA{\IEEEauthorrefmark{2}Michigan Technological University}
\IEEEauthorblockA{\IEEEauthorrefmark{3}University of Houston}
\IEEEauthorblockA{\{shiweid,lanzhang\}@mtu.edu,                 mpan2@uh.edu, xyyuan@mtu.edu}
}

% }

% \author{\IEEEauthorblockN{Shiwei Ding}
% \IEEEauthorblockA{\textit{College of Computing} \\
% \textit{Michigan Technology University}\\
% shiweid@mtu.edu}
% \and
% \IEEEauthorblockN{Lan Zhang}
% \IEEEauthorblockA{\textit{Department of Electrical and Computer Engineering} \\
% \textit{Michigan Technology University}\\
% lanzhang@mtu.edu}
% \and
% \IEEEauthorblockN{Miao Pan}
% \IEEEauthorblockA{\textit{Department of Electrical and Computer Engineering} \\
% \textit{University of Houston}\\
% mpan2@uh.edu}
% \and
% \IEEEauthorblockN{Xiaoyong Yuan}
% \IEEEauthorblockA{\textit{College of Computing} \\
% \textit{Michigan Technology University}\\
% xyyuan@mtu.edu}}

\maketitle

\begin{abstract}
   Collaborative inference has been a promising solution to enable resource-constrained edge devices to perform inference using state-of-the-art deep neural networks (DNNs). 
   In collaborative inference, the edge device first feeds the input to a partial DNN locally and then uploads the intermediate result to the cloud to complete the inference. However, recent research indicates model inversion attacks (MIAs) can reconstruct input data from intermediate results, posing serious privacy concerns for collaborative inference. Existing perturbation and cryptography techniques are inefficient and unreliable in defending against MIAs while performing accurate inference. This paper provides a viable solution, named \defense, which develops privacy-oriented pruning to balance privacy, efficiency, and utility of collaborative inference.
   \defense takes advantage of the fact that later layers in a DNN can extract more task-specific features. Given limited local resources for collaborative inference, \defense intends to deploy more layers at the edge based on pruning techniques to enforce task-specific features for inference and reduce task-irrelevant but sensitive features for privacy preservation. To achieve privacy-oriented pruning, \defense introduces two key components: Lipschitz regularization and adversarial reconstruction training, which increase the reconstruction errors by reducing the stability of MIAs and enhance the target inference model by adversarial training, respectively. On a real-world collaborative inference task, vehicle re-identification, we demonstrate the superior performance of \defense in terms of against MIAs.
\end{abstract}

\section{Introduction}
\label{sec:intro}

Collaborative inference has become a promising solution for using computationally intensive and memory-expensive state-of-the-art deep neural networks (DNNs) on resource-constrained edge devices.~\cite{he2019model,ko2018edge,teerapittayanon2017distributed}. 
In collaborative inference, a large-size DNN is divided into two partitions and deployed at the edge and the cloud. The input data observed at the edge is fed to the first DNN partition locally; the intermediate output is then sent to the cloud and processed remotely by the second DNN partition. The cloud eventually returns the inference result to edge devices. Collaborative inference can potentially serve a wide range of applications, offering great advantages over the conventional edge or cloud-only inference~\cite{wang2020convergence}.

\begin{figure}[!tb]
    \centering
\begin{subfigure}[t]{0.24\linewidth}
  % include first image
  \centering
  \includegraphics[width=\linewidth]{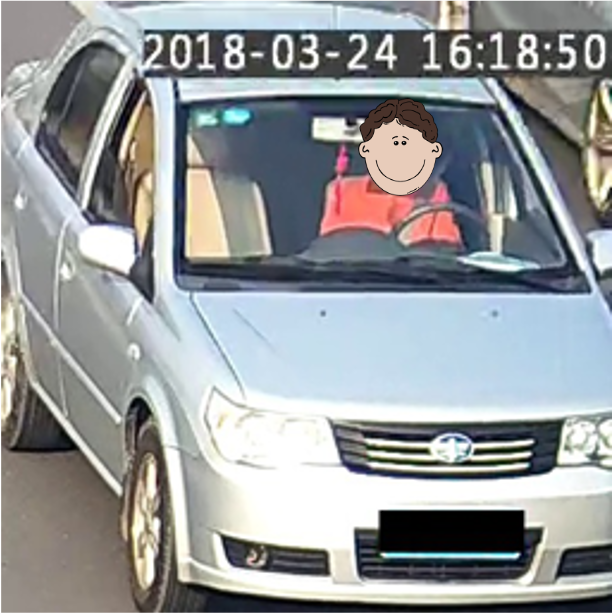}
   \caption{}%Original image
\end{subfigure}
\begin{subfigure}[t]{0.24\linewidth}
  % include first image
  \centering
  \includegraphics[width=\linewidth]{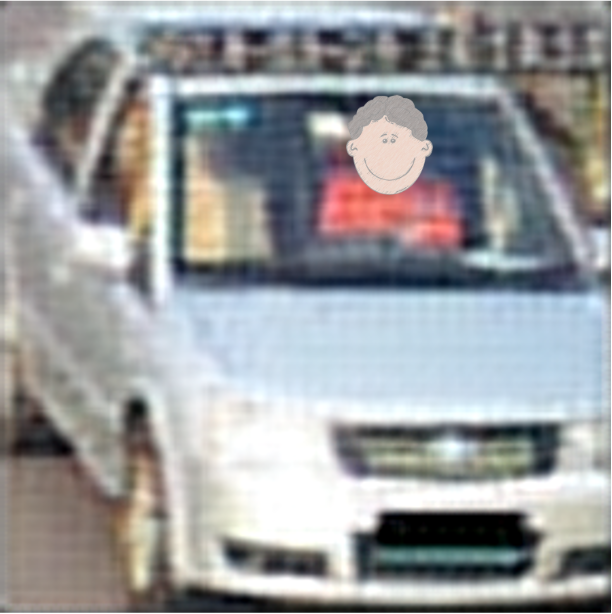}
   \caption{} %Reconstructed image without defenses}
\end{subfigure}
\begin{subfigure}[t]{0.24\linewidth}
  % include first image
  \centering
  \includegraphics[width=\linewidth]{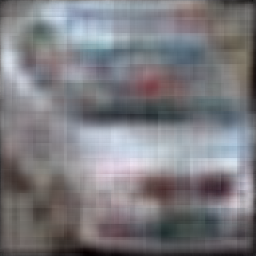}
   \caption{}
\end{subfigure}
\begin{subfigure}[t]{0.24\linewidth}
  % include first image
  \centering
  \includegraphics[width=\linewidth]{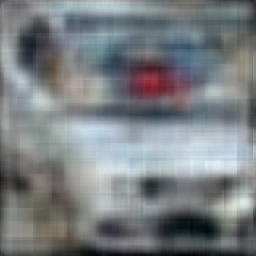}
   \caption{}
\end{subfigure}
\vspace{-0.5em}
\caption{Model inversion attacks (MIAs) under collaborative inference. (a) shows the original input image captured by a surveillance camera. (b) and (c) show reconstructed images by MIAs when two and four ResNet blocks are deployed at the edge, respectively. (d) shows the reconstructed image by MIA under \defense protection, where four pruned ResNet blocks are deployed at the edge. The four pruned ResNet blocks have a comparable number of parameters to the two original ResNet blocks. Real facial information is replaced to preserve privacy.
}
\label{fig:MIAdifferent_1}
\vspace{-1.5em}
\end{figure}

Nevertheless, recent research on model inversion attacks (MIAs) has identified privacy risks during collaborative inference~\cite{oh2019exploring,he2020attacking,xiao2020adversarial, pasquini2021unleashing,chen2020practical}. MIAs aim to reconstruct the confidential information of input data from intermediate results during inference~\cite{fredrikson2014privacy,zhang2020secret}. Due to the unknown communication environments or the untrusted cloud server, MIAs can observe intermediate outputs during collaborative inference and reconstruct raw inputs~\cite{hitaj2017deep,melis2019exploiting}, 
raising serious privacy concerns. Two mainstream defenses have been investigated against MIAs for privacy-preserving collaborative inference. First, perturbation techniques are used to modify intermediate outputs, such as by injecting noises to raw inputs, using adversarial training to add perturbations, or by randomly dropping out intermediate results~\cite{he2020attacking,xiao2020adversarial, ng2022ninjadesc,oh2019exploring}. Despite the computational efficiency, perturbation-based approaches sacrifice inference performance, such as accuracy, to a large extent for sufficient privacy guarantees. Second, cryptographic techniques ($e.g.$, secure multi-party computation~\cite{wagh2021falcon,mishra2020delphi} and homomorphic encryption~\cite{juvekar2018gazelle,liu2017oblivious}) are used to encrypt intermediate outputs before sharing. The cloud server will process the encrypted intermediate output without decryption. Although cryptographic techniques provide a strong privacy guarantee, they introduce significant delay and computational costs, such as a 14,000$\times$ slowdown~\cite{mireshghallah2021not}, rendering them ineffective for inference at resource-constrained edge devices, especially for the time-critical tasks. 

To address these limitations of existing defenses, in this paper, we propose a viable solution by developing privacy-oriented pruning, named \defense, to trade off privacy, efficiency, and utility of collaborative inference. \defense leverages the fact that the latter layer can extract more task-specific features from the input data than the previous layer of a DNN. When the entire DNN is deployed at the edge, the intermediate output becomes the inference result, potentially restricting MIAs' success. We illustrate this idea using the vehicle re-identification task as an example, where surveillance cameras upload intermediate results of captured images after local processing, based on which the cloud server identifies the same vehicle within these images. MIAs intend to reconstruct the task-irrelevant but sensitive information, $i.e.$, the driver's identity. We empirically evaluate MIA performance in Figure~\ref{fig:MIAdifferent_1}, where the overall inference uses a ResNet-18~\cite{he2016deep} with four ResNet blocks, followed by fully-connected layers. When two ResNet blocks are deployed at the edge, the driver’s identity can be revealed through intermediate results, as shown in (b). In comparison, it is difficult to infer the driver’s identity when four ResNet blocks are deployed at the edge, as shown in (c). Motivated by these observations, \defense intends to deploy more layers at the edge to enforce task-specific intermediate features for inference while reducing task-irrelevant but sensitive features for privacy preservation. However, given limited resource budgets at the edge, a critical question is how to store more layers cost-effectively while maintaining collaborative inference accuracy.

Neural network pruning has been well-recognized to determine the sub-network of a DNN to speed up inference without significantly sacrificing prediction performance~\cite{mozer1989skeletonization,han2015deep,blalock2020state}.
However, existing research mainly focuses on balancing the accuracy and efficiency of pruned models, where redundant model parameters in terms of accuracy are pruned rather than privacy-sensitive ones. Moreover, recent research indicates pruned models potentially have more serious privacy risks~\cite{yuan2022membership}.
Therefore, to carefully determine privacy-oriented pruning, \defense introduces two key components: Lipschitz regularization and adversarial reconstruction training. Enlighten by the effectiveness of \textit{Lipschitz regularization}  to improve the stability of DNNs~\cite{virmaux2018lipschitz,fazlyab2019efficient,cranko2021generalised}, \defense enforces a Lipschitz constraint in an opposite manner during pruning. When selecting a pruned network structure, the Lipschitz constraint increases the reconstruction error by reducing the stability of MIAs.
Besides, \defense employs the \textit{adversarial reconstruction training} to alternatively train the surrogate attacker and defender in a game fashion, which further strengthens the target model against strong MIAs.
We illustrate the reconstructed image of the vehicle re-identification task protected by \defense in Figure~\ref{fig:MIAdifferent_1}(c). Using a comparable number of edge-side parameters as in (b), \defense realizes the driver's identity protection in (d), which is as effective as deploying the entire model at the edge in (c).  
The major contributions of this paper are summarized below.
\begin{itemize}
\vspace{0.5em}
\item We develop \defense to defend against MIAs under collaborative inference based on privacy-oriented pruning. \defense trades off privacy, efficiency, and utility of collaborative inference, allowing resource-constrained edge devices using state-of-the-art DNNs for privacy-preserving inference. 
\vspace{0.5em}
\item We introduce two key components to \defense, the Lipschitz regularization and adversarial reconstruction training, which enable privacy-oriented pruning by enforcing task-specific features at intermediate outputs for accurate inference while reducing task-irrelevant but sensitive features for privacy preservation.
\vspace{0.5em}
\item We evaluate \defense on a real-world collaborative inference task, vehicle re-identification. \defense can compress the edge DNN partition by $66.7\%$ with only $3.1\%$ prediction accuracy loss on VeriWild datasets and compress edge DNN partition by around $92\%$ with $12.7\%$ prediction accuracy loss on Veri datasets. Meanwhile, \defense successfully reduces the MIAs performance by $11.9\%$, and $10.9\%$ in terms of two attack metrics, PSNR and SSIM, on the VeriWild dataset. Also, it decreases MIAs PSNR and SSIM performance by $21.5\%$ and $20.1\%$ on the VERI dataset. The results on both datasets demonstrate superior performance compared to the baseline defenses. 
\end{itemize}
\vspace{0.5em}

%------------------------------------------------------------------------
\vspace{-0.1em}
\section{Related Work}
\vspace{0.2em}
\subsection{Model Inversion Attacks}
\vspace{0.2em}
Model inversion attacks (MIAs) reconstruct confidential information of the raw input from a target model's outputs or intermediate results during the inference phase~\cite{fredrikson2014privacy,zhang2020secret}. 
MIAs were originally proposed to recover confidential information from training data~\cite{fredrikson2015model}, but recent research has shown that they also pose a threat to raw input data during inference.
For example, Yang et al.~\cite{yang2019neural} proposed an inversion network for input reconstruction, where the adversary feeds the target model's output into the inversion network and trains the inversion network to predict raw input data. 

Recently, MIAs have received attention under collaborative inference between the edge and the cloud platforms. Oh et al.~\cite{oh2019exploring} and He et al.~\cite{he2020attacking} conducted MIAs by reconstructing the input data from the intermediate results transferred from edge to cloud. 
Salem et al.~\cite{salem2020updates} and Pasquini et al.~\cite{pasquini2021unleashing} recently reconstructed the input data by developing an autoencoder and a generative adversarial network (GAN), respectively.
Recent research has revealed the significant threat posed by MIAs to collaborative inference.

%-------------------------------------------------------------------------
\vspace{0.2em}
\subsection{MIA Defenses under Collaborative Inference}
\vspace{0.2em}
To provide privacy-preserving collaborative inference, defenses against MIAs can be broadly categorized into cryptographic-based and perturbation-based approaches. Cryptographic technologies, such as homomorphic encryption (HE) and secure multi-party computation (SMC), have been widely used to protect inference data privacy~\cite{gilad2016cryptonets, liu2017oblivious}. 
Although offering strong privacy guarantees, cryptography-based defenses impose significant computational and communication overheads~\cite{mo2021ppfl}, making them infeasible for resource-constrained edge devices. 

Perturbation-based approaches have been another widely used MIA defense under collaborative inference. For example, Mireshghallah et al.~\cite{mireshghallah2021not} proposed to add noise to the features that do not contribute to the final inference result. He~\cite{he2020attacking} and Oh~\cite{oh2019exploring} randomly drop part of outputs and skip connection in the neural network. However, these methods significantly reduce the prediction performance. Another common way adding perturbation to defend against MIAs is the adversarial training method~\cite{edwards2015censoring, raval2017protecting, huang2017context, wu2018towards, pittaluga2019learning, wu2020privacy, xiao2020adversarial, ng2022ninjadesc, dave2022spact}. It tries to make the target model's outputs show less information about the inputs, which increases the difficulties for the adversary in reconstructing the user's inputs and maintaining the performance of the target model.
However, these methods often rely on large perturbations to achieve satisfactory protection against MIAs, which can significantly reduce prediction performance. To address the limitations of existing defenses, this paper aims to develop a more efficient approach to protect collaborative inference privacy against MIAs while maintaining high accuracy.

%-------------------------------------------------------------------------

\vspace{0.2em}
\subsection{Neural Network Pruning}
\vspace{0.2em}
Neural network pruning aims to compress a DNN model and increase inference efficiency by removing redundant parameters~\cite{han2015deep,lin2019towards}. Pruning techniques effectively address the challenges of resource constraints on edge devices and improve inference speed. Han et al.~\cite{han2015deep} proposed to remove the model parameters with large magnitudes to increase model efficiency. 
Recent research has introduced the use of masks to provide a soft pruning approach to model parameters. Yang et al.~\cite{8787574} developed a method that adds a mask for each neuron or filter to prune according to the importance of the parameter.  
Li et al.~\cite{Li_2019_CVPR} proposed to add a binary mask to prune the weights whose masks are 0 while training.
Lin et al.~\cite{lin2019towards} trained a soft mask to identify the importance of specific structures ($e.g.$, blocks, branches, or channels) in a neural network and prune the less important structures based on their soft mask values.
Inspired by Lin's work, we design a privacy-oriented pruning method to prune privacy-sensitive convolution channels or blocks.

%-------------------------------------------------------------------------
\vspace{0.2em}
\section{Methodology}
\label{sec:Methodology}
\begin{figure}[!tb]
\centering
\begin{subfigure}[b]{\linewidth}
  \centering
  \includegraphics[width=\linewidth]{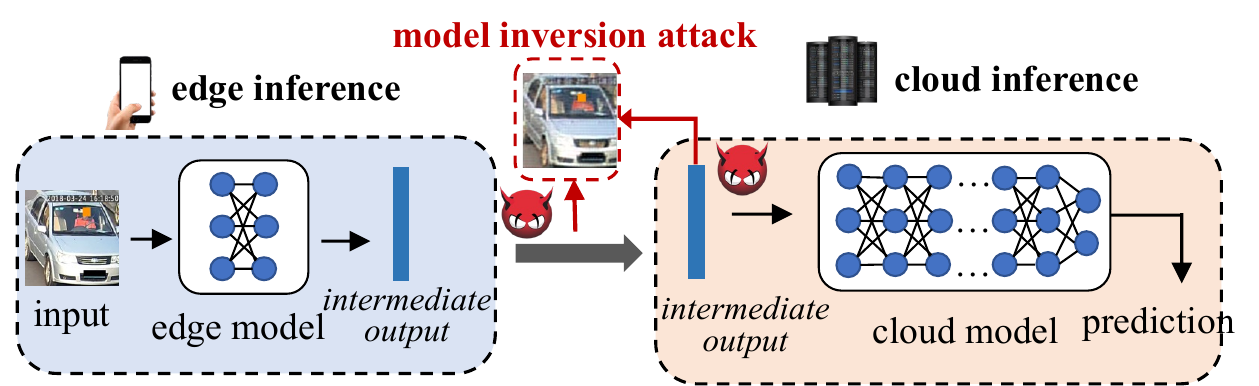}
  \caption{Original collaborative inference.}
%   \vspace{0.5em}
\end{subfigure}
\vspace{-0.5em}
\begin{subfigure}[b]{\linewidth}
  \centering
  \includegraphics[width=\linewidth]{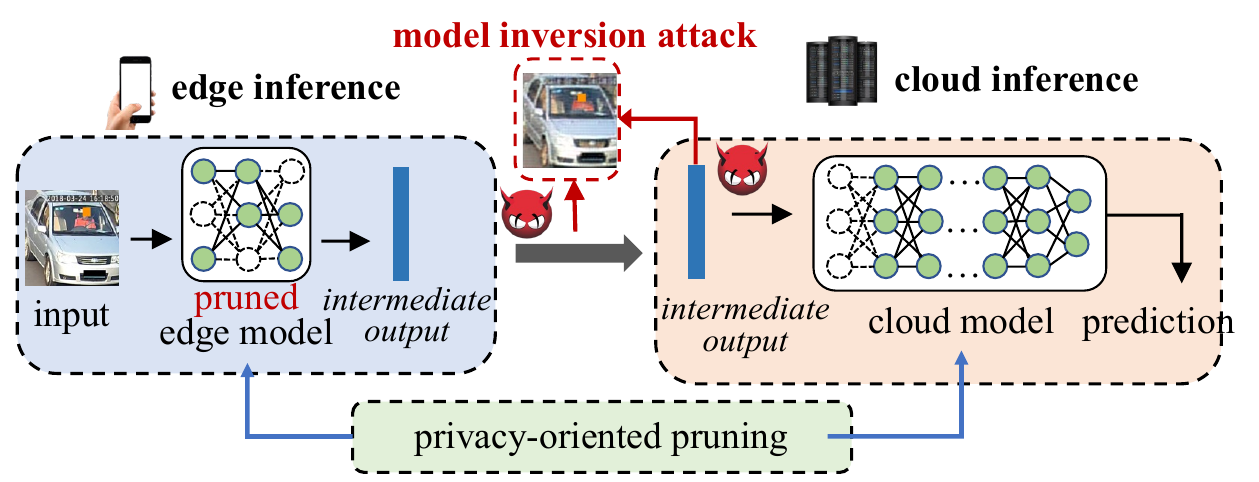}
\caption{Privacy-preserving collaborative inference via \defense.}
\end{subfigure}
\caption{Model inversion attacks against (a) original collaborative inference  and  (b) privacy-preserving collaborative inference protected by \defense.}
\vspace{-0.5em}
\label{fig:MIAprocess}
\vspace{-0.5em}
\end{figure}

This section introduces the methodology of \defense. We first present the threat model and then introduce the design of privacy-oriented pruning with two key components in \defense. Figure~\ref{fig:MIAprocess} illustrates original collaborative inference under MIAs in (a) and the proposed privacy-preserving collaborative inference protected by \defense in (b).

%-------------------------------------------------------------------------
\subsection{Threat Model}
\label{sec:Threat Model}
We consider a collaborative inference system between the edge and the cloud. 
The adversary aims to reconstruct the raw input from the intermediate output of the edge-side model as shown in Figure~\ref{fig:MIAprocess}. 
We assume the adversary has no access to the cloud and edge side model parameters (i.e., black-box MIA), but the adversary has prior knowledge about the dimension of the raw input, the architecture of the cloud and edge side models, and the training dataset.

Our design focuses on a practical collaborative inference scenario where edge devices have limited computing, communication, and storage resources. Thus, it is infeasible for the edge to store and process the entire DNN in its local memory or encrypt the inference data and upload it to the cloud in real-time.

%-------------------------------------------------------------------------
\vspace{0.2em}
\subsection{Privacy-Oriented Pruning}
\label{sec:Pruning reason}
\subsubsection{Design Overview}
\defense aims to protect the collaborative inference system against MIAs while preserving inference utility and efficiency. Our idea comes from the fact that the latter layer can extract more task-specific features from the input data than the previous layer of a DNN and reduce the task-irrelevant but sensitive features. 
To accommodate more neural network layers on the edge side for privacy preservation, we introduce a privacy-oriented neural network pruning that reduces the neural network size on the edge while maintaining the utility and efficiency of the edge model. 
We adopt structured pruning~\cite{anwar2017structured} in this paper, which can remove specific structures (e.g., channels or blocks) from a target model. Compared with unstructured pruning, structured pruning enables the hardware acceleration of edge devices for sparse matrix computation and accelerates the inference process. 

Define the well-trained DNN for collaborative inference by $f=f_c\circ f_e$, named as the target model. The cloud-side partition is defined by $f_c$ with parameters $\theta_c$, and the edge-side partition is defined by $f_e$ with parameters $\theta_e$. $\theta = \{\theta_e, \theta_c\}$ is trained on a training dataset $\mathcal{D}$. To remove structures, such as channels or blocks, from the target model $f$, we introduce a trainable soft mask $m$ in \defense training to scale the output of structures.
In channel-wise pruning, $m$ is applied to each channel's output.
In block-wise pruning, $m$ is applied to the residual mapping of each residual block.
A small value in the well-trained soft mask $m$ indicates that the output of its corresponding structure has little contribution to the final prediction. Removing these structures will not affect the prediction performance.
Therefore, we formulate the following optimization problem to train soft mask $m$ and target model $f$ with parameters $\theta$:
\begin{equation}
\min_{m} \mathcal{L}(\theta, m) + \lambda \|m\|_1 + \lambda_2 \|m\|_2, \vspace{-0.5em}
\label{eq:mask1}
\end{equation} 
where $\mathcal{L}(\theta, m)$ denotes the prediction loss (e.g., the cross-entropy loss in the classification tasks), $\|m\|_1$ and $\|m\|_2$ denote two sparsity regularizers, and $\lambda_1$ and $\lambda_2$ denote the hyper-parameters to balance the prediction loss and the sparsity regularizers.
We introduce $\ell_1$ sparsity regularizer ($\|m\|_1$) to reduce the number of structures in the target model and achieve a high sparsity ratio.
We also introduce $\ell_2$ sparsity regularizer ($\|m\|_2$) to control the magnitude of soft mask $m$.
Therefore, incorporating $\ell_2$ sparsity regularizer can reduce the regularization error caused by the $\ell_1$ regularizer to compromise between the high sparsity ratio and the prediction accuracy.
To achieve a high convergence rate and a high sparsity, we adopt a fast iterative shrinkage-thresholding (FISTA) algorithm~\cite{beck2009fast} to update the soft mask $m$.
Once the soft mask $m$ and target model $f$ have been trained, \defense prunes the channels or blocks of the target model if their corresponding soft mask values are smaller than a threshold $\tau$. In other words, the structures with little contribution to the model prediction are removed.

Existing pruning techniques have primarily focused on improving accuracy and efficiency rather than addressing privacy concerns. To protect data privacy using pruning, we further incorporate Lipschitz regularization and adversarial reconstruction training into the pruning process (detailed in Section~\ref{sec:Adversary reason} and~\ref{sec:Lipschitz reason}).
Lipschitz regularization aims to increase their reconstruction errors by reducing the stability of MIAs.
Adversarial reconstruction training generates a surrogate attacker during the training process to mislead the attacker in a game theoretic formulation.

\begin{algorithm}[]
\SetAlgoLined
\SetKwInOut{Input}{Input}\SetKwInOut{Output}{Output}
\caption{\defense}
\label{algo:algo1}
\Input{Training dataset $\mathcal{D}$, cloud-side model $f_c$ with parameters $\theta_c$, edge-side model $f_e$ with parameters $\theta_e$, entire DNN model $f=f_c\circ f_e$ with parameters $\theta = \{\theta_c, \theta_e\}$, total number of layers $N$, soft mask $m$ for pruning, max training epoch $T$, pruning threshold $\tau$, surrogate inversion model $f_{adv}$ with parameters $\theta_{adv}$.}
\Output{Pruned model $f_p$.}
Initialize the soft mask $m$ from a Normal distribution $m\sim \mathcal{N}(0, 1)$\\
Initialize the cloud-side, edge-side and surrogate inversion models $f_c$, $f_e$, $f_{adv}$\\
\For{epoch $t=1, \ldots, T$}{
    \If{t \% 10 = 0}{
        \% Train surrogate inversion model\\
        \For{batch sample $(x, y) \in \mathcal{D}$}{
            Reconstruct the raw input data: $x_{adv} = f_{adv}(f_e(x, \theta_e), \theta_{adv})$\\
            Update $\theta_{adv}$ to minimize $\mathcal{L}_{adv}$ in Eq.~\ref{eq:adv} \\
        }
    }

    \For{batch sample $(x, y) \in \mathcal{D}$}{
        \% Perform adversarial reconstruction training\\ 
        Reconstruct the raw input data: $x_{adv} = f_{adv}(f_e(x, \theta_e), \theta_{adv})$\\
                
        Update parameters $\theta$ to minimize the prediction loss $\mathcal{L}$ and maximize the adversarial loss $\mathcal{L}_{adv}$ in Eq.~\ref{eq:adv_loss}:\\ $\min_{\theta}\quad \mathcal{L}(\theta, m) - \beta \mathcal{L}_{adv}(\theta_e, \theta_{adv})$\\
        
        \BlankLine
        \% Perform Lipschitz regularization \\
        \For{block $i = 1, \ldots, N$}{
            Sample $\bm{\delta} \sim \mathcal{N}(0, 1)$\\
            Calculate Lipschitz constraint $k$ of the $i$-th block defined in Eq.~\ref{eq:Lip2} and~\ref{eq:Lip2_1}
        }
        Update parameters $\theta$ to minimize the prediction loss $\mathcal{L}$ and Lipschitz loss $\mathcal{L}_{lip}$ in Eq.~\ref{eq:lip}:\\
        $\min_{\theta}\quad \mathcal{L}(\theta, m) + \mathcal{L}_{lip}(\theta_e)$\\
        
        \BlankLine
        \% Train the soft mask $m$\\
        Update mask $m$ to minimize the prediction loss $\mathcal{L}$ and the size of $m$ using Eq.~\ref{eq:mask1}:\\
        $\min_{\theta, m}\quad \mathcal{L}(\theta, m) + \lambda_1 \|m\|_1 + \lambda_2 \|m\|_2$
    }
}
\% Perform privacy-oriented pruning\\
Derive pruned model $f_p$ by removing the channels or blocks if the corresponding mask $m_i \leq \tau$\\
\textbf{Return} Pruned model $f_p$ with parameter $\theta$.

\end{algorithm}
%-------------------------------------------------------------------------
\vspace{0.2em}
\subsubsection{Lipschitz Regularization}
\label{sec:Lipschitz reason}
We introduce Lipschitz regularization in pruning to reduce the stability of model inversion attacks and increase their reconstruction errors.
The idea of Lipschitz regularization is to restrict the changes of model output given a small input change, so that the output of DNN models will be stable given perturbations in the input~\cite{virmaux2018lipschitz,fazlyab2019efficient,cranko2021generalised}. We leverage Lipschitz regularization in an inverse fashion. We aim to make the model inversion attack unstable and increase the reconstruction errors by enforcing Lipschitz constraints.

Given a function $f$, the Lipschitz constant $k$ of $f$ is defined as the smallest constant in the Lipschitz condition:
\begin{equation}
\label{eq:Lip1}
k = \sup_{x_1\neq x_2} \frac{\|f(x_1) - f(x_2)\|}{\|x_1-x_2\|}. \vspace{-0.5em}
\end{equation}
Given a certain distance between outputs, the lower bound of the distance between inputs can be derived using Lipschitz constant $k$:
\begin{equation}
\|x_1-x_2\| \geq \frac{1}{k}\|f(x_1) - f(x_2)\|.
\end{equation}
The increase of $\frac{1}{k}$ will lead to a large difference in the reconstructed data, given a small difference in the output. 
Thus, we can defend against model inversion attacks by maximizing $1/k$ or minimizing the Lipschitz constant $k$. 

Since calculating the Lipschitz constant $k$ is an intractable problem~\cite{virmaux2018lipschitz}, we introduce the block-wise local Lipschitz constant $k_i$ to approximate the Lipschitz constant. 
Given an edge model $f_e$ with $N$ blocks, we denote $f_i$ as the $i$-th block of the model. $f = f_N\circ f_{N-1} \circ, \cdots f_1, (i=1,2, \cdots, N)$.
For $i\geq 2$, we define block-wise local Lipschitz constraint $k_i$ of the $i$-th block as:
\begin{equation}
k_{i} = \sup_{x}\frac{\|f_i f_{i-1} ... f_1(x+ \delta) - f_i f_{i-1} ... f_1(x)\|_1}{\|f_{i-1} ... f_1(x+ \delta) - f_{i-1}... f_1(x)\|_1},
\label{eq:Lip2}
\end{equation}
where $\delta$ denotes a random noise sampled from a Gaussian distribution.
For $i=1$, we define block-wise local Lipschitz constraint $k_1$ of the first block as:
\begin{equation}
k_{1} = \sup_{x} \frac{\|f_1(x+ {\delta}) - f_1(x)\|_1}{{\|\delta\|_1}}.\vspace{-0.5em}
\label{eq:Lip2_1}
\end{equation}

We calculate the Lipschitz loss using the block-wise local Lipschitz constraint as follows:

\begin{equation}
\label{eq:lip}
\mathcal{L}_{lip}(\theta_e) = \sum_{i=1}^{N} \alpha_i k_i,\vspace{-0.5em}
\end{equation}
where $\alpha_i$ is the hyper-parameter to balance the constraints. By minimizing the Lipschitz loss, we increase the accumulated errors of model inversion attacks over blocks.
We include the Lipschitz loss as a regularization term in the loss function and train the model parameters $\theta$ to minimize the prediction loss and the Lipschitz loss:
\begin{equation}
\label{eq:Lip3}
\mathrm{min}_{\theta}\quad \mathcal{L}(\theta, m) + \mathcal{L}_{lip}(\theta_e).\vspace{-0.5em}
\end{equation}

\vspace{0.2em}
\subsubsection{Adversarial Reconstruction Training}
\label{sec:Adversary reason}
We leverage adversarial reconstruction training to mislead the model inversion attacker and protect input data privacy. Specifically, we first generate a surrogate inversion model $f_{adv}$ with parameters $\theta_{adv}$.  Given an input sample $x$, the surrogate inversion model $f_{adv}$ aims to extract the raw input data from the intermediate output $f_e(x,\theta_e)$. The parameters $\theta_{adv}$ are trained to minimize the adversarial loss $\mathcal{L}_{adv}$, which measures the difference between the reconstructed data $f_{adv}(f_e(x, \theta_e)$ and raw input sample $x$. The adversarial loss $\mathcal{L}_{adv}$ can be calculated as:
\begin{equation}
\label{eq:adv}
\mathcal{L}_{adv}(\theta_e, \theta_{adv}) = \|x - f_{adv}(f_e(x, \theta_e), \theta_{adv})\|_2.
\end{equation}

By integrating the surrogate inversion model, the target model $f$ is trained to mislead the model inversion attackers while maintaining the prediction performance. To achieve this, we maximize the adversarial loss while minimizing the prediction loss by solving the optimization problem:
\begin{equation}
\label{eq:adv_loss}
\min_{\theta} \mathcal{L}(\theta, m) - \beta\mathcal{L}_{adv}(\theta_e, \theta_{adv}).\vspace{-0.5em}
\end{equation}

We aim to identify the strongest attack given a target model and incorporate the strongest attack into the minimization problem, which can be formulated as a bi-level optimization problem:
\setlength{\belowdisplayskip}{0.5em}
\begin{equation}
\min_{\theta} \max_{\theta_{adv}} \mathcal{L}(\theta, m) - \beta\mathcal{L}_{adv}(\theta_e, \theta_{adv}),\vspace{-0.5em}
\end{equation}
where the inner maximization problem is to find the strongest attack for the target model, and the outer minimization problem is to train a model to mislead the strongest attack.
Since it is computationally intensive to solve a bi-level optimization problem, in this paper, we train the target model parameters $\theta$ and the surrogate inversion model parameters $\theta_{adv}$ iteratively, following the common practice in adversarial reconstruction training.

%-------------------------------------------------------------------------

%-------------------------------------------------------------------------
\section{Evaluation}
This section presents ablation studies to show the effectiveness of the proposed designs in \defense. 
\vspace{0.2em}
\subsection{Experimental Settings}
\noindent
\textbf{Vehicle Re-identification Dataset.}
We consider a real-world collaborative inference task, vehicle re-identification, for evaluation. The vehicle re-identification task requires collaboration between multiple edge devices (e.g., surveillance cameras) and a cloud server. Each edge device processes the captured image and uploads the intermediate results to the cloud server. The cloud server identifies if two images capture the same vehicle.
Our experiments are conducted on two real-world vehicle re-identification datasets, VERIWild~\cite{lou2019large} and VERI~\cite{liu2016deep, liu2017provid}. VERIWild is the most recent and largest dataset for vehicle re-identification, capturing 416,314 images of 40,671 vehicles' identities from a large CCTV system with 174 cameras during one month. The VERI dataset contains 49,357 images of 776 vehicles from 20 cameras in 24 hours. 
We evaluate \defense on three test datasets with different sizes in VERIWild: small, medium, and large and VERI testing dataset. The small, medium, and large testing dataset in VERIWild contains 3,000, 5,000, and 10,000 vehicle identification, and 38,861, 64,389, and 128,517 images for testing, respectively. The VERI testing dataset contains 11579 images for testing.

\begin{figure}[!h]
    \centering
    \includegraphics[width=0.99\linewidth]{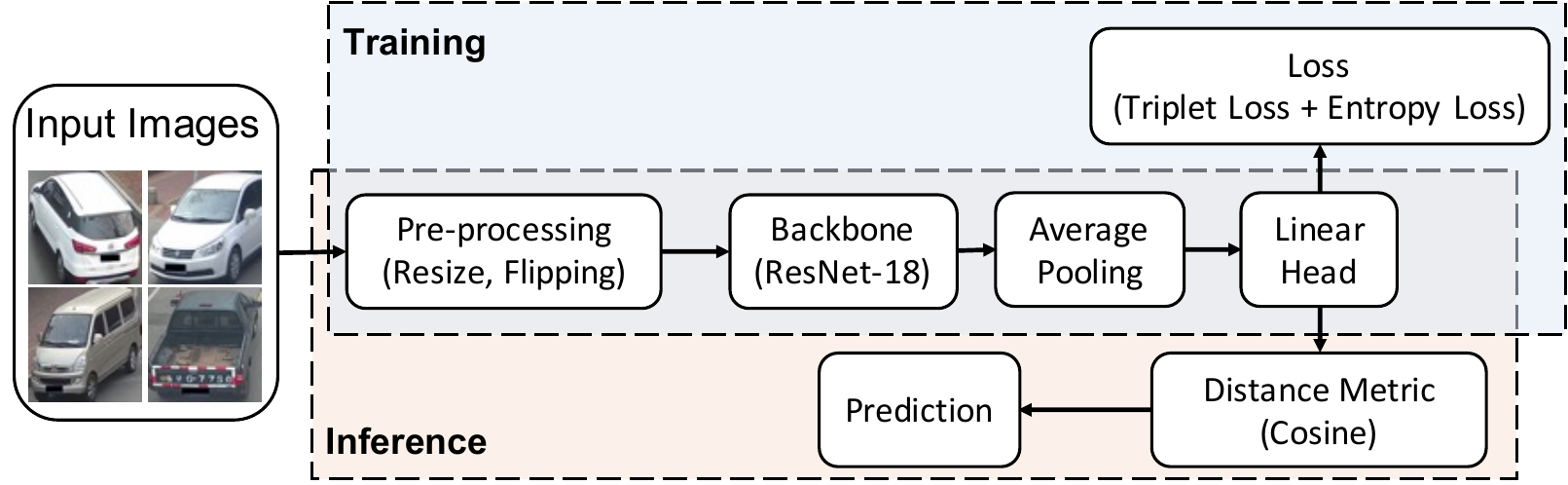}
    \caption{Vehicle re-identification framework used in the experiments.}
    \label{fig: Experiment Model}
\end{figure}

\vspace{0.2em}
\noindent
\textbf{Vehicle Re-identification Model.} We deploy ResNet-18~\cite{he2016deep} as the backbone model for the collaborative vehicle re-identification task and train the model using the open-source framework, Fastreid\footnote{fastreid, \url{https://github.com/JDAI-CV/fast-reid}}~\cite{he2020fastreid}.
Due to resource constraints, it is infeasible to deploy the entire ResNet-18 model on edge. In our work, we consider resource-constrained edge devices, where only the first one or two ResNet residual blocks can be deployed on edge (ResNet-18 includes $4$ residual blocks and $17$ convolution layers in total.).

\begin{table}[!tb]    
\centering
    \small
    \resizebox{\linewidth}{!}{%    
    \begin{tabular}{ c|rrrr}
        \toprule
        \centering
        Dataset & \multicolumn{1}{c}{Prediction Acc. Drop} & \multicolumn{1}{c}{PSNR Drop} & \multicolumn{1}{c}{SSIM Drop} & \multicolumn{1}{c}{Attack Acc. Drop} \\
        \midrule
        VERIWild Small & $2.0\%$ & $11.9\%$ & $11.7\%$ & $15.8\%$\\
        VERIWild Medium & $2.2\%$ & $12.0\%$ & $11.7\%$ & $25.7\%$\\
        VERIWild Large & $3.1\%$ & $11.9\%$ & $10.9\%$ & $28.0\%$\\
        VERI & $12.7\%$ & $21.5\%$ & $20.1\%$ & $14.5\%$\\
        \bottomrule
    \end{tabular}
    }
    \vspace{-0.5em}
    \caption{The defense performance of \defense on the small, medium, and large VERIWild datasets and VERI dataset. We run five different hyper-parameter settings in \defense and report the average values of Re-identification accuracy drop (Prediction Acc. Drop), PSNR drop, SSIM drop, and attack accuracy drop from the original target model without defenses. 
    }
    \label{tab: Experiment result average}
    \vspace{-1.5em}
\end{table}

\begin{figure}[!tb]
    \centering
\begin{subfigure}[t]{0.32\linewidth}
  \centering
  \includegraphics[width=.95\linewidth]{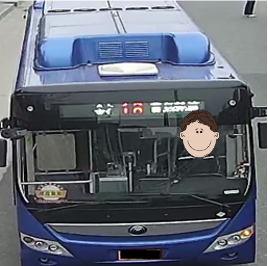}
\end{subfigure}
\begin{subfigure}[t]{0.32\linewidth}
  \centering
  \includegraphics[width=.95\linewidth]{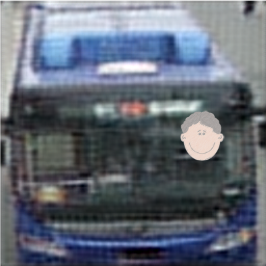}
\end{subfigure}
\begin{subfigure}[t]{0.32\linewidth}
  \centering
  \includegraphics[width=.95\linewidth]{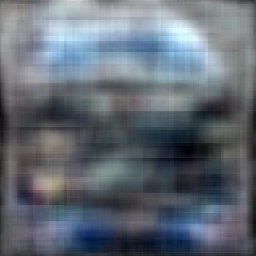}
\end{subfigure}

\vspace{0.2em}
\begin{subfigure}[t]{0.32\linewidth}
  \centering
  \includegraphics[width=.95\linewidth]{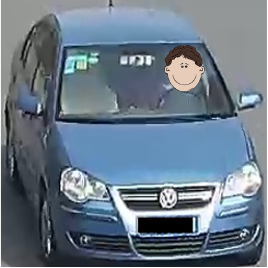}
\end{subfigure}
\begin{subfigure}[t]{0.32\linewidth}
  \centering
  \includegraphics[width=.95\linewidth]{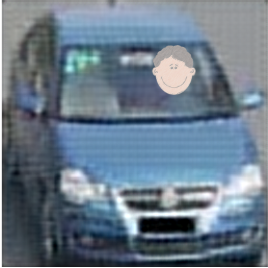}
\end{subfigure}
\begin{subfigure}[t]{0.32\linewidth}
  \centering
  \includegraphics[width=.95\linewidth]{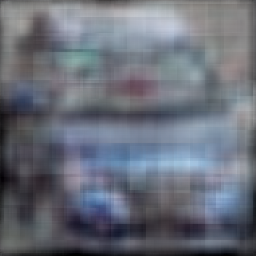}
\end{subfigure}

\vspace{-0.5em}
\caption{\defense performance. (a) shows the original images captured by the surveillance cameras. (b) shows the reconstructed images from the output of two ResNet blocks without defenses. Confidential information (e.g., vehicle brand, driver identity, vehicle interiors) is exposed in the images. (c) shows the reconstructed images from the output of four ResNet blocks that are protected by \defense. The confidential information becomes invisible after \defense. Real facial information is replaced to preserve privacy.}
\label{fig:MIAdifferent}
\vspace{-1em}
\end{figure}

\vspace{0.2em}
\noindent
\textbf{Experiments Hyper-parameter Settings}
We first train the target model for $90$ epochs using a learning rate of $0.0003$. 
Then we perform the proposed defense. In the proposed defense, we retrain the model for $90$ epochs with a learning rate $0.0003$ for the model and $10^{-6}$ for the soft mask.
We set batch size as $512$ and the threshold $\theta$ = $0$ for pruning.
We set hyper-parameters $\beta = 0.0004$, $\lambda_1 = 1$, $\lambda_2 = 10$, $\alpha_1 = 5$, $\alpha_2 = 0.2$, $\alpha_3 = 0.01$, $\alpha_4 = 0.005$.

\noindent
\textbf{The Data Process Details}
\label{sec:Exp setting}
Figure~\ref{fig: Experiment Model} illustrates the vehicle re-identification framework used for training and inference. 
We first pre-process the input data by resizing and random flipping. Then we deploy ResNet-18~\cite{he2016deep} as the backbone. 
The features are aggregated via average pooling. We choose a linear layer as the prediction head.
In the training phase, we include triplet loss and cross-entropy loss in the loss function and train the target model.
In the inference phase, given two input images, we calculate the cosine similarity between two predicted embeddings and make predictions.

\noindent
\textbf{Inversion Model and Surrogate Inversion Model.}
We implement a black-box inversion model following~\cite{he2020attacking}.
The inversion model is designed to invert the neural network layers of the target network. 
For instance, given a target neural network with three $3\times3$ convolution layers at the edge side, we deploy the inversion model with three $3\times3$ deconvolution layers. 
This design ensures that the inversion model aligns with the target neural network. 

Furthermore, we introduce a surrogate inversion model in adversarial regularization training.
Different from the inversion model, the surrogate inversion model uses the output of the last convolution layer of the target network as its input. We design the surrogate inversion model to invert all the neural network layers in the target network.

\vspace{0.2em}
\noindent
\textbf{Evaluation Metrics.}
We consider the three types of evaluation metrics:

 1) {\textit{Prediction Accuracy}}: We report the target model's top-1 accuracy on the test dataset to measure the model utility.
 
 2) \textit{PSNR} and \textit{SSIM}: We use the similarity between the original and reconstructed images to measure the privacy risks. Two commonly used similarity metrics, PSNR and SSIM~\cite{hore2010image}, are reported in the paper. The higher PSNR/SSIM indicates a higher reconstruction quality or worse defense performance. 

 3) \textit{L2 norm distance}: We select the average l2 distance between the input images and reconstruction images as an auxiliary metric. A lower L2 norm distance proves a higher privacy risk. 

 4) \textit{Attack accuracy}: We deploy a ResNet-50 model pre-trained on the ImageNet-1K dataset to predict the class of the reconstructed images. We select 13 categories related to vehicles in ImageNet-1K as target labels. If a reconstruction image has been categorized into the target labels, it means the attacker launched a successful attack. The accuracy for this classification model is called the attack accuracy.
 
\vspace{0.2em}
\noindent\textbf{Existing Defenses.}
Since our design focuses on resource-constrained edge devices, we compare \defense with three perturbation-based MIA defenses: 

(1) \textit{noise defense}, where the  intermediate output is perturbed by Gaussian noise~\cite{he2020attacking}.

(2) \textit{dropout defense}, where we randomly drop out the intermediate output~\cite{he2020attacking}.

(3) \textit{skip defense}, where we randomly skip connections between convolution layers~\cite{oh2019exploring}.

(4) \textit{Adversarial Privacy-Preserving}~\cite{xiao2020adversarial} (denoted by APP), which selects a reconstructed network and a discriminator to guarantee the reconstruction quality and use the reconstruction images for the adversarial training to force the output of the target network show less information about the input.
%-------------------------------------------------------------------------
\vspace{0.2em}
\subsection{Experimental Results}
\label{sec: Experimental Results}
\begin{figure}[!tb]
\centering
\begin{subfigure}[b]{\linewidth}
  \centering
  \includegraphics[width=0.99\linewidth]{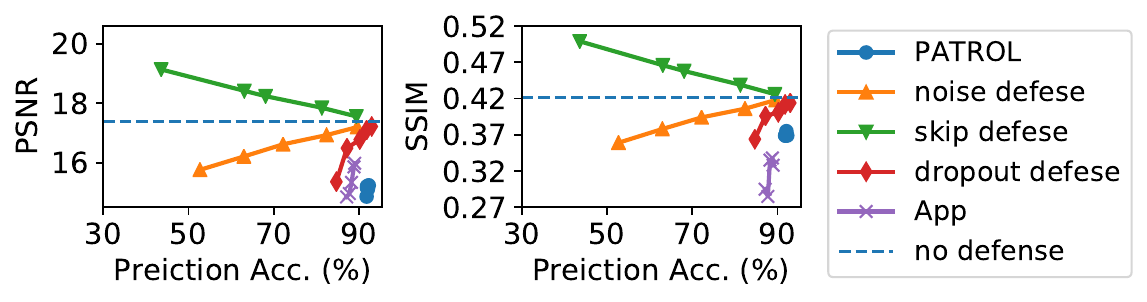}
  \caption{The reconstructed images results on VERIWild Small dataset.}
\end{subfigure}
\begin{subfigure}[b]{\linewidth}
  \centering
  \includegraphics[width=0.99\linewidth]{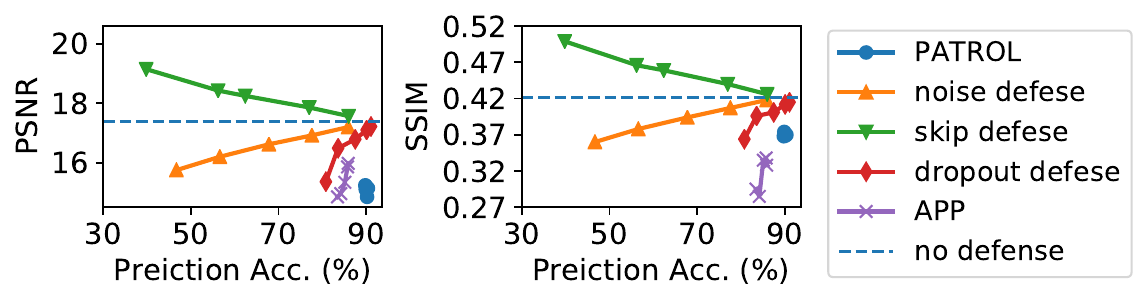}
  \caption{The reconstructed images results on VERIWild Medium dataset.}
\end{subfigure}
\begin{subfigure}[b]{\linewidth}
  \centering
  \includegraphics[width=0.99\linewidth]{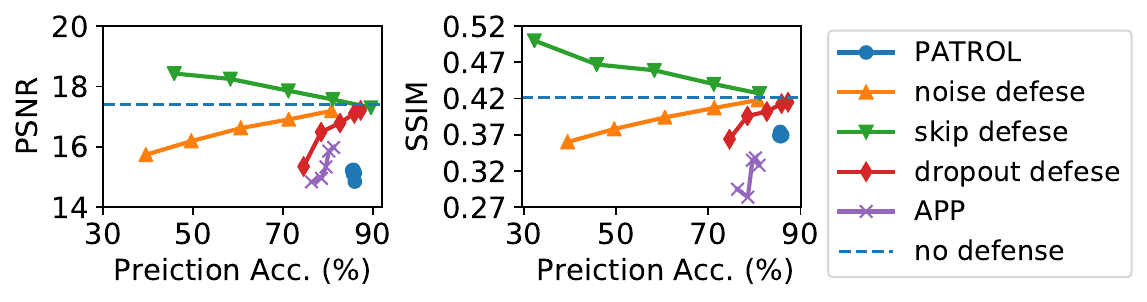}
  \caption{The reconstructed images results on VERIWild Large dataset.}
\end{subfigure}
\begin{subfigure}[b]{\linewidth}
  \centering
  \includegraphics[width=0.99\linewidth]{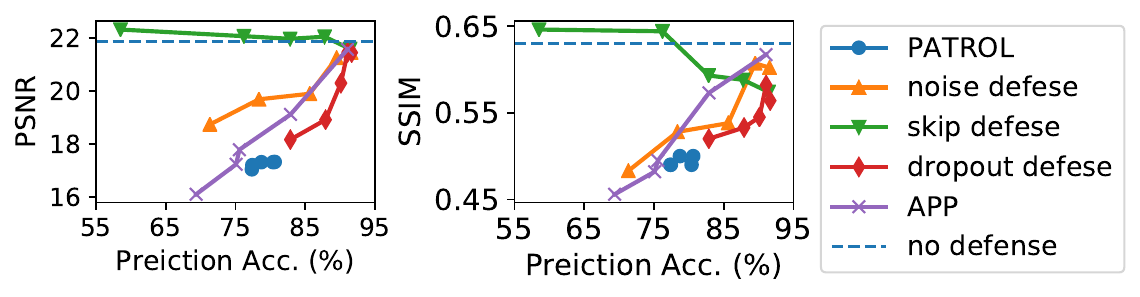}
    \caption{The reconstructed images results on VERI dataset.}
    \label{fig: VERI result}
\end{subfigure}
\caption{The PSNR and SSIM of reconstructed images on the VERIWild Small, Medium, Large dataset and VERI dataset. Low PSNR or SSIM indicates the low similarity between the reconstructed and original images, i.e., good protection performance. \defense outperforms the existing defenses, providing a better tradeoff between privacy and utility.}
\label{fig:PSNR AND SSIM}
\vspace{-1em}
\end{figure}

\noindent\textbf{Effectiveness of \defense}. 
We evaluate the performance of \defense on three test datasets and compare it with the original model without defenses.
We assume that edge devices can only deploy $2$ residual blocks (memory footprint $2.9$MB) in the original model without defenses for the VERIWild dataset and can deploy $1$ residual block (memory footprint $0.6$MB) in the original model without defenses for the VERI dataset.
By using \defense with a high pruning ratio that removes around $66.7\%$ parameters for the VERIWild dataset and around $92\%$ parameters for the VERI dataset, $4$ residual blocks ($4.8$MB memory cost for VERIWild and $0.5$MB memory cost for VERI dataset) can be deployed on edge (the impact of different pruning ratios will be discussed in Section~\ref{sec:Experiments and Evaluation}).
To present the effectiveness of \defense, we measure the prediction accuracy drop, PSNR drop, SSIM drop, and attack accuracy drop from the original model.
As shown in Table~\ref{tab: Experiment result average}, \textit{\defense significantly degrades the MIA attack performance in terms of PSNR, SSIM, and attack accuracy, while incurring an affordable prediction accuracy drop.}
More importantly, \defense effectively preserves confidential information, such as the driver's identity, in collaborative inference models, as shown in Figure~\ref{fig:MIAdifferent}.

\vspace{0.0em}
\noindent
\textbf{Comparison with Existing Defenses}.
We compare \defense with five existing defenses and the original model without defenses (baseline).
Figure~\ref{fig:PSNR AND SSIM} reports the model performance (prediction accuracy) and privacy risks (PSNR and SSIM), $i.e.$, the ineffectiveness of defenses.
The blue dashed line indicates the performance of the original model without any defenses.

As shown in Figure~\ref{fig:PSNR AND SSIM}, the proposed \defense achieves the lowest privacy risks compared with the noise, dropout, and skip connection defenses on two datasets. Under the same prediction accuracy, the PSNR and SSIM metrics for \defense drop at least $11\%$ and $10\%$ than these three defenses on VERIWild datasets and drop at least $2\%$ and $1\%$ on VERI dataset. Under the same privacy level, \defense prediction accuracy is higher at least $10\%$ than the three defense methods on the VERIWild dataset and at least $3\%$ on the VERI dataset.  
Compared with the adversarial training defense methods APP~\cite{xiao2020adversarial}, the \defense achieves a similar defense performance with higher prediction accuracy (around $4\%$) on the VERIWild and VERI dataset, which demonstrates \defense capability to achieve a better trade-off for the model accuracy and privacy than the APP.

Although noise defense and dropout defense achieve a slightly better prediction accuracy when we apply little noise or a small dropout ratio, their privacy protection becomes ineffective (high PSNR and SSIM values). 
As we add more noise or set a high dropout ratio, their defense performance can be the same as or better than \defense, but the prediction accuracy of the target model drops dramatically. 
The skip defense is ineffective in protecting the model's privacy, although it improves the model's efficiency when we skip more neurons in the network.
The APP method performs better than noise, dropout and skip defenses on VERIWild datasets. But on the VERI dataset, the App's advantage is not obvious compared with dropout and noise defense. The APP method achieves better prediction accuracy but less privacy-preserving when using a small trade-off parameter for adversarial training. When we apply a large trade-off parameter for the adversarial training, the APP method can keep the privacy but receives a low model accuracy.

In summary, noise defense and dropout defense methods fail to strike a balance between prediction accuracy and privacy protection performance. The APP method has a similar trade-off as \defense, but our method takes advantage of the APP methods in two datasets.
In contrast, \textit{\defense achieves both high model prediction accuracy and effective privacy protection}. Moreover, the pruning employed by \defense reduces the model size and maintains the efficiency of the edge model.

%-------------------------------------------------------------------------

\subsection{Ablation Study}
\label{sec:Experiments and Evaluation}
This section presents ablation studies to demonstrate the effectiveness of the proposed designs in \defense. We evaluate the results on VERIWild dataset for all experiments in this section.

%-------------------------------------------------------------------------
\vspace{0.2em}
\noindent
\textbf{Effectiveness of Lipschitz Regularization and Adversarial Reconstruction Training}.
\label{Sec: adv_lip}
Lipschitz regularization and adversarial reconstruction training are two key components to enable privacy-oriented pruning.
We investigate the effectiveness of Lipschitz regularization and adversarial reconstruction training.
We consider three different defense approaches for the ablation study.
First, only structured pruning is used for defense (pruning-only).
Second, we integrate the structure pruning with adversarial reconstruction training (pruning with Adv).
Third, we integrate the structure pruning with Lipschitz regularization (pruning with Lip).
We compare the performance of these three defense approaches with \defense.
As shown in Table~\ref{tab:3 Compare-3}, all the defense methods can degrade the attacker's performance. 
Both Lipschitz regularization and adversarial reconstruction training can reduce the privacy risks (large PSNR and SSIM drop) compared with the pruning-only defense.
By integrating both Lipschitz regularization and adversarial reconstruction training, the privacy-oriented pruning in \defense achieves the best performance in both accuracy (smallest accuracy drop, $3.2$\%) and privacy protection (largest PSNR and SSIM drop, $11.9$\% and $10.9$\%).

\begin{table}[!tb]    
    \centering
    \resizebox{\linewidth}{!}{
    \begin{tabular}{ c|rrr}
        \toprule
        Defenses & Accuracy Drop  & PSNR Drop & SSIM Drop\\ 
        \midrule
        Pruning-only & $5.4\%$ & $8.3\%$ & $7.1\%$\\
        Pruning with Adv & $5.0\%$ & $10.3\%$ & $10.9\%$\\
        Pruning with Lip & $5.7\%$ & $9.5\%$ & $9.5\%$\\
        \textbf{\defense} & \textbf{3.2\%} & \textbf{11.9\%} & \textbf{10.9\%}\\
        \bottomrule
    \end{tabular}
    }
    \vspace{-0.5em}
    \caption{Effectiveness of \defense components. Both adversarial reconstruction training (Adv) and Lipschitz Regularization (Lip) reduce the attack performance compared to the pruning-only approach.
    }
    \label{tab:3 Compare-3}
    \vspace{-0.5em}
\end{table}

\begin{table}[!tb]
    \centering
    \resizebox{\linewidth}{9.6mm}{
    \begin{tabular}{ c|rrr}
        \toprule
        Pruning Method & Accuracy Drop & PSNR Drop & SSIM Drop\\ 
        \midrule
        \textbf{Channel-wise} & \textbf{3.1\%} & \textbf{11.9\%} & \textbf{10.9\%}\\
        Block-wise & $10.5\%$ & $9.2\%$ & $9.5\%$\\
        Dropout defense& $10.4\%$ & $3.9\%$ & $7.1\%$\\
        \bottomrule
    \end{tabular}}
    \vspace{-0.5em}
    \caption{
    Comparison of \defense using channel-wise and block-wise pruning. Channel-wise pruning achieves better defense performance and higher prediction accuracy. 
    Both pruning methods in \defense outperform dropout defense (the best defense baseline). 
    }
    \label{tab:Channel-Block}
    \vspace{-1em}
\end{table}
%-------------------------------------------------------------------------
\begin{table}[!tb]
\centering
\resizebox{\linewidth}{!}{%
\begin{tabular}{ c|rrr} 
 \toprule
  Defenses & \begin{tabular}[c]{@{}c@{}}\# of blocks \\ on edge\end{tabular} & PSNR Drop& SSIM Drop\\
 \midrule
  No defense & 1 & 0\% & 0\% \\
  \defense w/o pruning & $1$ & $0\%$ & $0\%$\\
  \defense w/ low pruning ratio & $2$ & $4.6\%$ & $8.9\%$\\
  \defense w/ high pruning ratio & $3$ & \textbf{10.2\%} & \textbf{15.6\%}\\
 \bottomrule
\end{tabular}
}
\caption{\defense Performance on small devices. Only one residual block can be deployed on the edge device without pruning (no defense and \defense without pruning). \defense with a low pruning ratio can deploy two residual blocks on the edge device, while \defense with a high pruning ratio can deploy three residual blocks. 
}
\label{tab:First-case Result}
\vspace{-0.5em}
\end{table}

\begin{table}[!tb]
\centering
\resizebox{\linewidth}{!}{%
\begin{tabular}{ c|rrr} 
 \toprule
  Defenses & \begin{tabular}[c]{@{}c@{}}\# of blocks \\ on edge\end{tabular} & PSNR Drop& SSIM Drop\\
 \midrule
  No defense & 2 & 0\% & 0\% \\
  \defense w/o pruning & $2$ & $1.8\%$ & $7.1\%$\\
  \defense w/ low pruning ratio & $3$ & $5.1\%$ & $9.5\%$\\
  \textbf{\defense w/ high pruning ratio} & $4$ & \textbf{11.2\%}& \textbf{11.9\%}\\
 \bottomrule
\end{tabular}
}
\caption{\defense Performance on large devices. 
Only two residual blocks can be deployed on the edge device without pruning. 
\defense with a low pruning ratio can deploy three residual blocks on the edge device, while \defense with a high pruning ratio can deploy four residual blocks. 
}
\label{tab:Second-case Result}
\vspace{-1em}
\end{table}

\vspace{0.2em}
\noindent
\textbf{Effectiveness of Pruning.}
We investigate the effectiveness of the pruning ratio in defense. We consider  both low and high pruning ratios and investigate two scenarios by taking edge device capabilities into account. In the small edge device scenario, only one residual block in ResNet-18 can be deployed on the edge, due to the limited memory. After privacy-oriented pruning, two or three blocks can be deployed on the edge, based on the low/high pruning ratio.
In the large edge device scenario, two residual blocks can be deployed on the edge side. After privacy-oriented pruning, three or four blocks can be deployed on the edge, based on the low/high pruning ratio.

Table~\ref{tab:First-case Result} and Table~\ref{tab:Second-case Result} illustrate the results in the two scenarios, where four settings are considered: 1) no defense, 2) \defense without pruning, which only uses Lipschitz regularization and Adversarial reconstruction training, 3) \defense with low pruning ratio, and 4) \defense with high pruning ratio. 
The reconstructing images' PNSR and SSIM from the original model without defense in the first scenario (low pruning ratio) is $18.15$ and $0.45$, and in the second scenario (high pruning ratio) is $17.16$ and $0.42$. 
We observe that the proposed pruning method achieves lower PSNR and SSIM values, providing much better privacy protection than the model without pruning, i.e., no defense and \defense without pruning.
As the high pruning ratio makes more layers to be deployed on the device, \defense with a high pruning ratio can achieve better performance than that with a low pruning ratio.
However, we find that when the edge device deploys more layers for the edge-side model, the privacy protection effectiveness of the pruning method decreases. In the Small Edge Device Scenario, \defense with a low pruning ratio, deploying one more residual block than the original model on the edge device, can reduce the PNSR and SSIM value by $0.84$ and $0.04$ when excluding the effect of the adversarial reconstruction training and Lipschitz regularization (Compared to the defense result between \defense with a low pruning ratio and the model with Adversarial and Lipschitz defense only). However, in the Large Edge Device Scenario, when excluding the effect of adversarial reconstruction training and Lipschitz regularization and deploying one more residual block deploying on the edge device, \defense with a low pruning ratio could only reduce the PNSR and SSIM by $0.56$ and $0.01$, the same observation also appears in every comparison model. The observation shows that the privacy protection of pruning method can reduce more PSNR and SSIM when fewer layers are on the edge-side model before pruning, which indicates the defensive pruning method is more effective on an edge device with small memory.

\vspace{0.2em}
\noindent
\textbf{Effectiveness of Pruned Model Structures.}
\label{sec:pruning flex}
In the previous experiments, we prune the target network by removing the channel of each convolution block, i.e., channel-wise pruning. 
Here, we consider another structure pruning method, block-wise pruning, where the entire convolution block can be removed from the target network. In our experiment, we add the soft mask at the end of each basic convolution block of ResNet-18 to implement block-wise pruning.
Table~\ref{tab:Channel-Block} demonstrates that the channel-wise pruning method yields higher model accuracy and better defense performance compared to the block-wise pruning method. 
This is mainly due to the trainable masks. The block-wise only has $8$ trainable masks (There are only $8$ convolution blocks in ResNet-18), which is hard to balance the trade-off between model accuracy and defense performance after pruning.
Despite its limitations, block-wise pruning has demonstrated some advantages over existing defenses. 
In light of the strong defense performance of the dropout defense among the existing defenses, we have included it in the table for comparison.

\begin{table}[!h]    
    \small
    \centering
    \vspace{-0.3em}
    \begin{tabular}{@{}llrr@{}}
    \toprule
    Defense & Attack & PSNR & SSIM \\ \midrule
    \multirow{2}{*}{No defense} & Black-box & 17.18 & 0.43 \\
     & White-box & 20.48 & 0.53 \\\hline
    \multirow{2}{*}{PATROL} & Black-box & 14.85 & 0.37 \\
     & White-box & 14.39 & 0.30 \\ \bottomrule
    \end{tabular}
    \vspace{-0.3em}
    \caption{Attack performance under black-box and white-box attacks with and without PATROL defense.}
    \label{tab: White-box attacking average}
    \vspace{-0.3em}
\end{table}

\noindent\textbf{Defense against White-Box Model Inversion Attacks.} The aforementioned evaluation mainly considers black-box attacks due to their popularity, while the proposed PATROL can effectively defend against white-box attacks. Table~\ref{tab: White-box attacking average} compares the black-box attacks and white-box attacks, where white-box attackers know the models' parameters (e.g., via reverse engineering). We observe that without defense, the white-box attack achieves higher attack performance (higher PSNR and SSIM) compared to the black-box attack on the model without any defense. However, \textit{by deploying PATROL, the attack performance of the white-box attacks is comparable or even lower than that of the black-box attacks.} This is mainly because the Lipschitz regularization used in \defense is attack-agnostic, which does not specifically target any particular attacks, which indicates that PATROL is effective against various types of attacks, including both white-box and black-box attacks.
\begin{figure}[!h]
    %[width=\textwidth]
    \centering
    \includegraphics[width=0.85\linewidth]{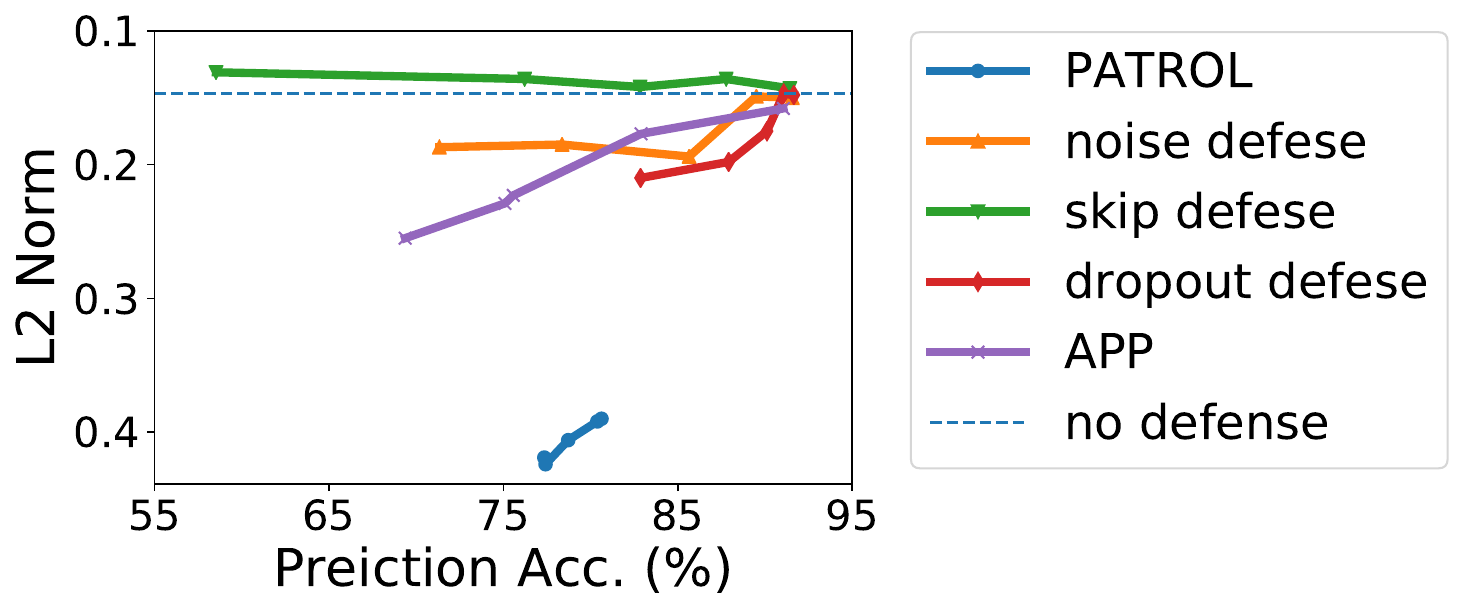}
    \vspace{-0.5em}
    \caption{The MSE (L2 norm) between reconstruction images and input images for \defense and the Baseline methods on the VERI dataset.}
    \label{fig: L2 Norm}
    \vspace{-0.6em}
\end{figure}
\vspace{-0.2em}

\noindent\textbf{The L2 Norm Metric Comparison Between \defense and Baseline Methods.} 
\label{sec: L2norm}
In this section, we consider using the Mean Square Error for the reconstruction image and the input image as an auxiliary metric to present the privacy-preserving for every defense method. In this experiment, we only use the VERI dataset as the testing dataset. The results are shown in Figure~\ref{fig: L2 Norm}.

The results show the same trend for each defense method as the PSNR and SSIM metrics in Figure~\ref{fig: VERI result}. The difference is that \defense performance is far beyond any other methods, which can be additional proof of the effectiveness of \defense. 
\vspace{-0.5em}

\begin{table}[!tb]    
\centering
    \small
    % \raggedright
    \resizebox{\linewidth}{!}{    
    \begin{tabular}{ c|rrrr}
        \toprule
        Defenses & PSNR Drop & SSIM Drop & Prediction Acc. Drop \\
        \midrule
        No defense & $3.0\%$ & $19.7\%$ & $22.8\%$\\
        APP~\cite{xiao2020adversarial} & $9.1\%$ & $33.8\%$ & $23.5\%$\\
        \defense & $6.2\%$ & $24.6\%$ & $18.3\%$\\
        \bottomrule
    \end{tabular}
    }
    \vspace{-0.5em}
    \caption{The privacy metric and Re-identification prediction accuracy drop (Prediction Acc. Drop) when we change the input size image form $256 \times 256$ to $128 \times 128$. We use the Adversarial privacy-preserving method (APP) as a comparison method.
    }
    \label{tab: Different input size impact}
    \vspace{-2.0em}
\end{table}

\noindent\textbf{The Impact of Image Size to the Defense.}
We try to figure out whether the input image size of vehicle identification influences the defense against MIAs. We use \defense, APP, and no defense model as the testing method, choosing $256 \times 256$ (default settings for all experiments) and $128 \times 128$ as two input sizes for this experiment.
Table~\ref{tab: Different input size impact} presents the evaluation metrics changes when the input size changes from $256 \times 256$ to $128 \times 128$ for no defense, APP and \defense methods. The results prove that the input size can influence the performance of the defense. The SSIM metrics have been more impacted by it since its metric value dropped more than the PSNR metric. But considering the prediction accuracy drop after changing the input size, the performance downgrade is caused by the downgrade of the target model performance (The no defense model prediction accuracy drops the same as the defense methods). Therefore, the input size has less impact on the defense against MIAs.
\begin{table}[!tb]    
\centering
    \small
    % \raggedright
    % \resizebox{\linewidth}{!}{%    
    \begin{tabular}{ c|rrr}
        \toprule
        \centering
        Dataset & \multicolumn{1}{c}{Defenses} & \multicolumn{1}{c}{Attack Accuracy}\\
        \midrule
        \multirow{6}{*}{VERIWild} & No Defense & $27.00\%$\\
         & APP~\cite{xiao2020adversarial} & $0.00\%$\\
         & Noise & $0.00\%$\\
         & Dropout & $0.00\%$\\
         & Skip connection & $0.00\%$\\
         & \defense & $0.00\%$\\
        \midrule
        \multirow{6}{*}{VERI} & No Defense & $14.50\%$\\
         & APP~\cite{xiao2020adversarial} & $0.00\%$\\
         & Noise & $0.00\%$\\
         & Dropout & $0.00\%$\\
         & Skip connection & $0.00\%$\\
         & \defense & $0.00\%$\\
        \bottomrule
    \end{tabular}
    % }
    \vspace{-0.5em}
    \caption{The attack accuracy for the classification model on no defense and defend models. The results of the VERIWild dataset are the average attack accuracy on the VERIWild small, medium, and large datasets. We confirm the attack accuracy metric is not meaningful for every model with defense.
    }
    \label{tab: Classification Results}
    \vspace{-1em}
\end{table}

\vspace{0.3em}
\noindent
\textbf{The Experiments Results for Classification Base Metrics of Different Defense Methods.}
\label{sec: Classification}
In table~\ref{tab: Classification Results}, we present the results of the prediction values for the classification network on different defense methods. If there is no defense for the target model, the attack accuracy (the accuracy of the classification model) reaches $26.00\%$ for VERIWild and $14.50\%$ for the VERI dataset. After we apply the defense methods, the attack accuracy for every defense method is $0\%$, which makes the comparison very difficult. Hence, we do not provide the attack accuracy of the classification model in the comparison of \defense and baselines.
%-----------------------------------------------------------------------
\section{Conclusion}
This paper proposed a privacy-oriented pruning for collaborative inference, named \defense, to defend against model inversion attacks.
\defense specially selects the sub-network of a DNN to push more layers to the edge without largely degrading prediction accuracy and efficiency. To remove the privacy-sensitive parameters, we introduced Lipschitz regularization and adversarial reconstruction training in \defense.
Defense performance of \defense is evaluated on vehicle re-identification tasks. Experimental results show that the proposed \defense can successfully protect collaborative inference against MIAs. 

\vspace{0.2em}
\noindent
\textbf{Acknowledgments:} This work is supported in part by the National Science Foundation under Grants CCF-2106754, CCF-2221741, and CNS-2151238.

{\small
\bibliographystyle{IEEEtran}
\bibliography{egbib, compress,deep, privacy}

\begin{thebibliography}{10}
\providecommand{\url}[1]{#1}
\csname url@samestyle\endcsname
\providecommand{\newblock}{\relax}
\providecommand{\bibinfo}[2]{#2}
\providecommand{\BIBentrySTDinterwordspacing}{\spaceskip=0pt\relax}
\providecommand{\BIBentryALTinterwordstretchfactor}{4}
\providecommand{\BIBentryALTinterwordspacing}{\spaceskip=\fontdimen2\font plus
\BIBentryALTinterwordstretchfactor\fontdimen3\font minus
  \fontdimen4\font\relax}
\providecommand{\BIBforeignlanguage}[2]{{%
\expandafter\ifx\csname l@#1\endcsname\relax
\typeout{** WARNING: IEEEtran.bst: No hyphenation pattern has been}%
\typeout{** loaded for the language `#1'. Using the pattern for}%
\typeout{** the default language instead.}%
\else
\language=\csname l@#1\endcsname
\fi
#2}}
\providecommand{\BIBdecl}{\relax}
\BIBdecl

\bibitem{he2019model}
Z.~He, T.~Zhang, and R.~B. Lee, ``Model inversion attacks against collaborative
  inference,'' in \emph{Proceedings of the 35th Annual Computer Security
  Applications Conference}, 2019, pp. 148--162.

\bibitem{ko2018edge}
J.~H. Ko, T.~Na, M.~F. Amir, and S.~Mukhopadhyay, ``Edge-host partitioning of
  deep neural networks with feature space encoding for resource-constrained
  internet-of-things platforms,'' in \emph{2018 15th IEEE International
  Conference on Advanced Video and Signal Based Surveillance (AVSS)}.\hskip 1em
  plus 0.5em minus 0.4em\relax IEEE, 2018, pp. 1--6.

\bibitem{teerapittayanon2017distributed}
S.~Teerapittayanon, B.~McDanel, and H.-T. Kung, ``Distributed deep neural
  networks over the cloud, the edge and end devices,'' in \emph{2017 IEEE 37th
  international conference on distributed computing systems (ICDCS)}.\hskip 1em
  plus 0.5em minus 0.4em\relax IEEE, 2017, pp. 328--339.

\bibitem{wang2020convergence}
X.~Wang, Y.~Han, V.~C. Leung, D.~Niyato, X.~Yan, and X.~Chen, ``Convergence of
  edge computing and deep learning: A comprehensive survey,'' \emph{IEEE
  Communications Surveys \& Tutorials}, vol.~22, no.~2, pp. 869--904, 2020.

\bibitem{oh2019exploring}
H.~Oh and Y.~Lee, ``Exploring image reconstruction attack in deep learning
  computation offloading,'' in \emph{The 3rd International Workshop on Deep
  Learning for Mobile Systems and Applications}, 2019, pp. 19--24.

\bibitem{he2020attacking}
Z.~He, T.~Zhang, and R.~B. Lee, ``Attacking and protecting data privacy in
  edge--cloud collaborative inference systems,'' \emph{IEEE Internet of Things
  Journal}, vol.~8, no.~12, pp. 9706--9716, 2020.

\bibitem{xiao2020adversarial}
T.~Xiao, Y.-H. Tsai, K.~Sohn, M.~Chandraker, and M.-H. Yang, ``Adversarial
  learning of privacy-preserving and task-oriented representations,'' in
  \emph{Proceedings of the AAAI Conference on Artificial Intelligence},
  vol.~34, no.~07, 2020, pp. 12\,434--12\,441.

\bibitem{pasquini2021unleashing}
D.~Pasquini, G.~Ateniese, and M.~Bernaschi, ``Unleashing the tiger: Inference
  attacks on split learning,'' in \emph{Proceedings of the 2021 ACM SIGSAC
  Conference on Computer and Communications Security}, 2021, pp. 2113--2129.

\bibitem{chen2020practical}
H.~Chen, H.~Li, G.~Dong, M.~Hao, G.~Xu, X.~Huang, and Z.~Liu, ``Practical
  membership inference attack against collaborative inference in industrial
  iot,'' \emph{IEEE Transactions on Industrial Informatics}, vol.~18, no.~1,
  pp. 477--487, 2020.

\bibitem{fredrikson2014privacy}
M.~Fredrikson, E.~Lantz, S.~Jha, S.~Lin, D.~Page, and T.~Ristenpart, ``Privacy
  in pharmacogenetics: An $\{$End-to-End$\}$ case study of personalized
  warfarin dosing,'' in \emph{23rd USENIX Security Symposium (USENIX Security
  14)}, 2014, pp. 17--32.

\bibitem{zhang2020secret}
Y.~Zhang, R.~Jia, H.~Pei, W.~Wang, B.~Li, and D.~Song, ``The secret revealer:
  Generative model-inversion attacks against deep neural networks,'' in
  \emph{Proceedings of the IEEE/CVF conference on computer vision and pattern
  recognition}, 2020, pp. 253--261.

\bibitem{hitaj2017deep}
B.~Hitaj, G.~Ateniese, and F.~Perez-Cruz, ``Deep models under the gan:
  information leakage from collaborative deep learning,'' in \emph{Proceedings
  of the 2017 ACM SIGSAC conference on computer and communications security},
  2017, pp. 603--618.

\bibitem{melis2019exploiting}
L.~Melis, C.~Song, E.~De~Cristofaro, and V.~Shmatikov, ``Exploiting unintended
  feature leakage in collaborative learning,'' in \emph{2019 IEEE symposium on
  security and privacy (SP)}.\hskip 1em plus 0.5em minus 0.4em\relax IEEE,
  2019, pp. 691--706.

\bibitem{ng2022ninjadesc}
T.~Ng, H.~J. Kim, V.~T. Lee, D.~DeTone, T.-Y. Yang, T.~Shen, E.~Ilg,
  V.~Balntas, K.~Mikolajczyk, and C.~Sweeney, ``Ninjadesc: content-concealing
  visual descriptors via adversarial learning,'' in \emph{Proceedings of the
  IEEE/CVF Conference on Computer Vision and Pattern Recognition}, 2022, pp.
  12\,797--12\,807.

\bibitem{wagh2021falcon}
S.~Wagh, S.~Tople, F.~Benhamouda, E.~Kushilevitz, P.~Mittal, and T.~Rabin,
  ``Falcon: Honest-majority maliciously secure framework for private deep
  learning,'' \emph{Proceedings on Privacy Enhancing Technologies}, vol.~1, pp.
  188--208, 2021.

\bibitem{mishra2020delphi}
P.~Mishra, R.~Lehmkuhl, A.~Srinivasan, W.~Zheng, and R.~A. Popa, ``Delphi: A
  cryptographic inference service for neural networks,'' in \emph{29th USENIX
  Security Symposium (USENIX Security 20)}, 2020, pp. 2505--2522.

\bibitem{juvekar2018gazelle}
C.~Juvekar, V.~Vaikuntanathan, and A.~Chandrakasan, ``$\{$GAZELLE$\}$: A low
  latency framework for secure neural network inference,'' in \emph{27th USENIX
  Security Symposium (USENIX Security 18)}, 2018, pp. 1651--1669.

\bibitem{liu2017oblivious}
J.~Liu, M.~Juuti, Y.~Lu, and N.~Asokan, ``Oblivious neural network predictions
  via minionn transformations,'' in \emph{Proceedings of the 2017 ACM SIGSAC
  conference on computer and communications security}, 2017, pp. 619--631.

\bibitem{mireshghallah2021not}
F.~Mireshghallah, M.~Taram, A.~Jalali, A.~T.~T. Elthakeb, D.~Tullsen, and
  H.~Esmaeilzadeh, ``Not all features are equal: Discovering essential features
  for preserving prediction privacy,'' in \emph{Proceedings of the Web
  Conference 2021}, 2021, pp. 669--680.

\bibitem{he2016deep}
K.~He, X.~Zhang, S.~Ren, and J.~Sun, ``Deep residual learning for image
  recognition,'' in \emph{Proceedings of the IEEE conference on computer vision
  and pattern recognition}, 2016, pp. 770--778.

\bibitem{mozer1989skeletonization}
M.~Mozer and P.~Smolensky, ``Skeletonization: {A} technique for trimming the
  fat from a network via relevance assessment,'' in \emph{{Advances in Neural
  Information Processing Systems (NIPS)}}, 1988.

\bibitem{han2015deep}
S.~Han, H.~Mao, and W.~J. Dally, ``Deep compression: Compressing deep neural
  network with pruning, trained quantization and huffman coding,'' in
  \emph{{International Conference on Learning Representations (ICLR)}}, 2016.

\bibitem{blalock2020state}
D.~W. Blalock, J.~J.~G. Ortiz, J.~Frankle, and J.~V. Guttag, ``What is the
  state of neural network pruning?'' in \emph{MLSys}, 2020.

\bibitem{yuan2022membership}
\BIBentryALTinterwordspacing
X.~Yuan and L.~Zhang, ``Membership inference attacks and defenses in neural
  network pruning,'' in \emph{31st USENIX Security Symposium (USENIX Security
  22)}.\hskip 1em plus 0.5em minus 0.4em\relax Boston, MA: USENIX Association,
  Aug. 2022, pp. 4561--4578. [Online]. Available:
  \url{https://www.usenix.org/conference/usenixsecurity22/presentation/yuan-xiaoyong}
\BIBentrySTDinterwordspacing

\bibitem{virmaux2018lipschitz}
A.~Virmaux and K.~Scaman, ``Lipschitz regularity of deep neural networks:
  analysis and efficient estimation,'' \emph{Advances in Neural Information
  Processing Systems}, vol.~31, 2018.

\bibitem{fazlyab2019efficient}
M.~Fazlyab, A.~Robey, H.~Hassani, M.~Morari, and G.~Pappas, ``Efficient and
  accurate estimation of lipschitz constants for deep neural networks,''
  \emph{Advances in Neural Information Processing Systems}, vol.~32, 2019.

\bibitem{cranko2021generalised}
Z.~Cranko, Z.~Shi, X.~Zhang, R.~Nock, and S.~Kornblith, ``Generalised lipschitz
  regularisation equals distributional robustness,'' in \emph{International
  Conference on Machine Learning}.\hskip 1em plus 0.5em minus 0.4em\relax PMLR,
  2021, pp. 2178--2188.

\bibitem{fredrikson2015model}
M.~Fredrikson, S.~Jha, and T.~Ristenpart, ``Model inversion attacks that
  exploit confidence information and basic countermeasures,'' in
  \emph{Proceedings of the 22nd ACM SIGSAC Conference on Computer and
  Communications Security}, 2015, pp. 1322--1333.

\bibitem{yang2019neural}
Z.~Yang, J.~Zhang, E.-C. Chang, and Z.~Liang, ``Neural network inversion in
  adversarial setting via background knowledge alignment,'' in
  \emph{Proceedings of the 2019 ACM SIGSAC Conference on Computer and
  Communications Security}, 2019, pp. 225--240.

\bibitem{salem2020updates}
A.~Salem, A.~Bhattacharya, M.~Backes, M.~Fritz, and Y.~Zhang,
  ``$\{$Updates-Leak$\}$: Data set inference and reconstruction attacks in
  online learning,'' in \emph{29th USENIX security symposium (USENIX Security
  20)}, 2020, pp. 1291--1308.

\bibitem{gilad2016cryptonets}
R.~Gilad-Bachrach, N.~Dowlin, K.~Laine, K.~Lauter, M.~Naehrig, and J.~Wernsing,
  ``Cryptonets: Applying neural networks to encrypted data with high throughput
  and accuracy,'' in \emph{International conference on machine learning}.\hskip
  1em plus 0.5em minus 0.4em\relax PMLR, 2016, pp. 201--210.

\bibitem{mo2021ppfl}
F.~Mo, H.~Haddadi, K.~Katevas, E.~Marin, D.~Perino, and N.~Kourtellis, ``Ppfl:
  privacy-preserving federated learning with trusted execution environments,''
  in \emph{Proceedings of the 19th Annual International Conference on Mobile
  Systems, Applications, and Services}, 2021, pp. 94--108.

\bibitem{edwards2015censoring}
H.~Edwards and A.~Storkey, ``Censoring representations with an adversary,''
  \emph{arXiv preprint arXiv:1511.05897}, 2015.

\bibitem{raval2017protecting}
N.~Raval, A.~Machanavajjhala, and L.~P. Cox, ``Protecting visual secrets using
  adversarial nets,'' in \emph{2017 IEEE Conference on Computer Vision and
  Pattern Recognition Workshops (CVPRW)}.\hskip 1em plus 0.5em minus
  0.4em\relax IEEE, 2017, pp. 1329--1332.

\bibitem{huang2017context}
C.~Huang, P.~Kairouz, X.~Chen, L.~Sankar, and R.~Rajagopal, ``Context-aware
  generative adversarial privacy,'' \emph{Entropy}, vol.~19, no.~12, p. 656,
  2017.

\bibitem{wu2018towards}
Z.~Wu, Z.~Wang, Z.~Wang, and H.~Jin, ``Towards privacy-preserving visual
  recognition via adversarial training: A pilot study,'' in \emph{Proceedings
  of the European conference on computer vision (ECCV)}, 2018, pp. 606--624.

\bibitem{pittaluga2019learning}
F.~Pittaluga, S.~Koppal, and A.~Chakrabarti, ``Learning privacy preserving
  encodings through adversarial training,'' in \emph{2019 IEEE Winter
  Conference on Applications of Computer Vision (WACV)}.\hskip 1em plus 0.5em
  minus 0.4em\relax IEEE, 2019, pp. 791--799.

\bibitem{wu2020privacy}
Z.~Wu, H.~Wang, Z.~Wang, H.~Jin, and Z.~Wang, ``Privacy-preserving deep action
  recognition: An adversarial learning framework and a new dataset,''
  \emph{IEEE Transactions on Pattern Analysis and Machine Intelligence},
  vol.~44, no.~4, pp. 2126--2139, 2020.

\bibitem{dave2022spact}
I.~R. Dave, C.~Chen, and M.~Shah, ``Spact: Self-supervised privacy preservation
  for action recognition,'' in \emph{Proceedings of the IEEE/CVF Conference on
  Computer Vision and Pattern Recognition}, 2022, pp. 20\,164--20\,173.

\bibitem{lin2019towards}
S.~Lin, R.~Ji, C.~Yan, B.~Zhang, L.~Cao, Q.~Ye, F.~Huang, and D.~Doermann,
  ``Towards optimal structured cnn pruning via generative adversarial
  learning,'' in \emph{Proceedings of the IEEE/CVF Conference on Computer
  Vision and Pattern Recognition}, 2019, pp. 2790--2799.

\bibitem{8787574}
C.~Yang, Z.~Yang, A.~M. Khattak, L.~Yang, W.~Zhang, W.~Gao, and M.~Wang,
  ``Structured pruning of convolutional neural networks via l1
  regularization,'' \emph{IEEE Access}, vol.~7, pp. 106\,385--106\,394, 2019.

\bibitem{Li_2019_CVPR}
T.~Li, B.~Wu, Y.~Yang, Y.~Fan, Y.~Zhang, and W.~Liu, ``Compressing
  convolutional neural networks via factorized convolutional filters,'' in
  \emph{Proceedings of the IEEE/CVF Conference on Computer Vision and Pattern
  Recognition (CVPR)}, June 2019.

\bibitem{anwar2017structured}
S.~Anwar, K.~Hwang, and W.~Sung, ``Structured pruning of deep convolutional
  neural networks,'' \emph{ACM Journal on Emerging Technologies in Computing
  Systems (JETC)}, vol.~13, no.~3, pp. 1--18, 2017.

\bibitem{beck2009fast}
A.~Beck and M.~Teboulle, ``A fast iterative shrinkage-thresholding algorithm
  for linear inverse problems,'' \emph{SIAM journal on imaging sciences},
  vol.~2, no.~1, pp. 183--202, 2009.

\bibitem{lou2019large}
Y.~Lou, Y.~Bai, J.~Liu, S.~Wang, and L.-Y. Duan, ``Veri-wild: A large dataset
  and a new method for vehicle re-identification in the wild,'' in
  \emph{Proceedings of the IEEE Conference on Computer Vision and Pattern
  Recognition}, 2019, pp. 3235--3243.

\bibitem{liu2016deep}
X.~Liu, W.~Liu, T.~Mei, and H.~Ma, ``A deep learning-based approach to
  progressive vehicle re-identification for urban surveillance,'' in
  \emph{Computer Vision--ECCV 2016: 14th European Conference, Amsterdam, The
  Netherlands, October 11-14, 2016, Proceedings, Part II 14}.\hskip 1em plus
  0.5em minus 0.4em\relax Springer, 2016, pp. 869--884.

\bibitem{liu2017provid}
------, ``Provid: Progressive and multimodal vehicle reidentification for
  large-scale urban surveillance,'' \emph{IEEE Transactions on Multimedia},
  vol.~20, no.~3, pp. 645--658, 2017.

\bibitem{he2020fastreid}
L.~He, X.~Liao, W.~Liu, X.~Liu, P.~Cheng, and T.~Mei, ``Fastreid: A pytorch
  toolbox for general instance re-identification,'' \emph{arXiv preprint
  arXiv:2006.02631}, 2020.

\bibitem{hore2010image}
A.~Hore and D.~Ziou, ``Image quality metrics: Psnr vs. ssim,'' in \emph{2010
  20th international conference on pattern recognition}.\hskip 1em plus 0.5em
  minus 0.4em\relax IEEE, 2010, pp. 2366--2369.

\end{thebibliography}
}

\end{document}